\documentclass[journal, twoside, print]{tmi_arxiv}
\usepackage{cite}
\usepackage{amsmath,amssymb,amsfonts}
\usepackage{mathtools}
\usepackage{algorithmic}
\usepackage{graphicx}
\usepackage{textcomp}
\usepackage{float}
\usepackage{tabularx}
\usepackage{multirow}
\usepackage{diagbox}
\usepackage{outlines}
\usepackage{fancyhdr}
\fancypagestyle{AllPages}{
\chead{\small This article has been accepted for publication in IEEE Transactions on Medical Imaging. \\ Citation information: DOI 10.1109/TMI.2022.3222126}
\rhead{\thepage}

}
\renewcommand{\textcolor}[2]{#2}

\usepackage{hyperref}

\begin{document}
\thispagestyle{empty}
\onecolumn
\begin{center}
\vspace*{4cm}
    \Large © 2022 IEEE.  Personal use of this material is permitted.  Permission from IEEE must be obtained for all other uses, in any current or future media, including reprinting/republishing this material for advertising or promotional purposes, creating new collective works, for resale or redistribution to servers or lists, or reuse of any copyrighted component of this work in other works.
\end{center}
\newpage
\twocolumn
\thispagestyle{empty}
\thispagestyle{AllPages}

\title{Federated Cycling (FedCy): Semi-supervised Federated Learning of Surgical Phases}
\author{Hasan Kassem, Deepak Alapatt, Pietro Mascagni, Consortium AI4SafeChole, Alexandros Karargyris, and Nicolas Padoy
\thanks{
Hasan Kassem and Deepak Alapatt contributed equally to this work.
 This work was partially supported by French State Funds managed by the “Agence Nationale de la Recherche (ANR)” through the “Investissements d’Avenir” (Investments for the Future) Program under Grant ANR-10-IAHU-02 (IHU-Strasbourg) and through the National AI Chair program under Grant ANR-20-CHIA-0029-01 (Chair AI4ORSafety). This work was granted access to the HPC resources of IDRIS under the allocation 2021-AD011011640R1 made by GENCI. }
 \thanks{The AI4SafeChole consortium is represented by Giovanni Guglielmo Laracca, Ludovica Guerriero, Andrea Spota, Claudio Fiorillo, Giuseppe Quero, Segio Alfieri, Ludovica Baldari, Elisa Cassinotti, Luigi Boni, Diego Cuccurullo, Guido Costamagna, and Bernard Dallemagne}
\thanks{Hasan Kassem, Deepak Alapatt, and Nicolas Padoy are affiliated with ICube, University of Strasbourg, CNRS, France
(email: \{h.kassem, alapatt, npadoy\}@unistra.fr)}
\thanks{Pietro Mascagni, Alexandros Karargyris, and Nicolas Padoy are affiliated with IHU Strasbourg, Strasbourg, France}
\thanks{Pietro Mascagni is affiliated with Fondazione Policlinico Universitario A. Agostino Gemelli IRCCS, Rome, Italy}}

\maketitle

\thispagestyle{empty}
\thispagestyle{AllPages}

\begin{abstract}

Recent advancements in deep learning methods bring computer-assistance a step closer to fulfilling promises of safer surgical procedures. However, the generalizability of such methods is often dependent on training on diverse datasets from multiple medical institutions, which is a restrictive requirement considering the sensitive nature of medical data. Recently proposed collaborative learning methods such as Federated Learning (FL) allow for training on remote datasets without the need to explicitly share data. Even so, data annotation still represents a bottleneck, particularly in medicine and surgery where clinical expertise is often required. With these constraints in mind, we propose FedCy, a federated semi-supervised learning (FSSL) method that combines FL and self-supervised learning to exploit a decentralized dataset of both labeled and unlabeled videos, thereby improving performance on the task of surgical phase recognition. By leveraging temporal patterns in the labeled data, FedCy helps guide unsupervised training on unlabeled data towards learning task-specific features for phase recognition. We demonstrate significant performance gains over state-of-the-art FSSL methods on the task of automatic recognition of surgical phases using a newly collected multi-institutional dataset of laparoscopic cholecystectomy videos. Furthermore, we demonstrate that our approach also learns more generalizable features when tested on data from an unseen domain.

\end{abstract}

\thispagestyle{empty}
\thispagestyle{AllPages}
\section{Introduction}
\label{sec:introduction}

\IEEEPARstart{O}{ver} the last decade, surgical data science\cite{maier2017surgical} has witnessed a surge of innovative applications due to breakthroughs in artificial intelligence and improvements in hardware acceleration. The rise of such applications can also be largely attributed to algorithms developed and benchmarked on large publicly available labeled datasets reflecting a high level of surgical expertise. While this approach has demonstrated significant value, indicative of the potential of surgical data science to disrupt clinical practice, concerns regarding the scalability and generalizability of these methods are frequently echoed through much of the literature in the field \cite{maier2022surgical}. Rightly, these concerns must be better understood, quantified, and addressed as we move towards the deployment of deep learning models in real-world settings.

Firstly, the scalability of label-intensive approaches is severely hindered by its dependence on scarce annotations that are manually and laboriously generated by overburdened clinical professionals\cite{maier2022surgical}. This bottleneck has limited the size of the few datasets that drive large chunks of research in the field. This has led to the development of methods to efficiently use fewer labels to supervise the training of deep learning models. In practice, this has meant using unlabeled data or easier-to-obtain labels in semi- or weakly- supervised settings\cite{yengera2018less,dipietro2019automated,yu2018learning,shi2021semi}, respectively. 

While significant progress has been made, inching towards fully-supervised performance using progressively fewer and fewer labels, potential biases towards the demographics and medical institutions represented in public datasets still remain. This is in part due to prohibitive restrictions preventing the free transfer and publishing of sensitive medical data due to privacy and medico-legal concerns. Recently, a decentralized learning technique known as federated learning\cite{pmlr-v54-mcmahan17a} has been gaining popularity allowing for the training of models on remote edge devices or servers, circumventing the need to explicitly exchange data. 

\subsection{Contributions}
\textcolor{blue}{
Our work falls squarely at the intersection of federated and semi-supervised learning. 
We present a method, FedCy, for federated semi-supervised learning of surgical phases, that allows learning from completely unlabeled datasets using a single completely labeled dataset by leveraging temporal patterns that occur in all. We focus this work on being able to scale to independently-sourced unlabeled datasets considering the practical value of being able to adapt deep learning models to other medical centers without the need to simultaneously scale annotation expertise, processes, and time. We briefly summarize our primary contributions below:
\begin{itemize}
    \item We believe FedCy is the first work applying federated learning to surgical videos and hope this work will serve as a foundation for other video-based tasks in the surgical domain.
    \item To the best of our knowledge, FedCy is the first work using distinct but complementary training objectives for labeled and unlabeled datasets. Using this setup, we achieve state-of-the-art results for federated semi-supervised learning on both labeled and unlabeled data for the task of surgical phase recognition.
    \item We conduct an extensive ablation quantifying the contribution of each included component and the value of our proposed configuration over both na\"ive and more sophisticated alternatives.
    \item We introduce a large, international multicenter dataset for surgical phase recognition containing 180 video recordings collected from 5 hospitals.
    \item We demonstrate that FedCy is more robust to out-of-distribution data than state-of-the-art approaches and shed light on this important research direction.
\end{itemize}
}

\section{Context}
\label{sec:context}
\textcolor{blue}{Surgical workflow modeling is often regarded as a key enabler to being able to analyze and support surgical procedures. In this regard, the ability to model a surgical procedure as a series of interlinked stages could enable systems that can describe, understand, explain, optimize, learn, teach, and eventually, automate the surgical process \cite{neumuth2017surgical}. Surgical phase recognition is the coarse recognition of surgical workflow and is often formulated as a classification task of video frames to one of several predefined phases. Over the last several years, surgical phase recognition has received growing attention from the surgical data science community \cite{garrow2021machine}, resulting in increasingly performant models \cite{twinanda2016endonet,czempiel2020tecno, gao2021trans} and clinically relevant applications such as automatic video documentation \cite{mascagni2021computer} and detection of deviations from the ``normal'' operative course \cite{berlet2022surgical}. Still,} these methods were neither trained nor evaluated on videos from multiple hospitals, limiting their deployment in the real world. This is mainly due to the sensitive nature of medical data that impedes data sharing and, by consequence, data aggregation and training on centralized datasets. Recently, less invasive alternative collaborative methods have been proposed for leveraging data from multiple sources in a privacy-preserving fashion.

\subsection{Federated Learning}

Federated Learning has emerged recently as a promising collaborative learning method\cite{ciil1} to mitigate data privacy concerns and the dependence of deep learning methods on developing in large centralized data lakes\cite{pmlr-v54-mcmahan17a}. In its vanilla setting, federated learning is a process where a central server coordinates and allows for multiple data owners to train collaboratively without having to explicitly expose their data. Data owners can be medical institutions or data centers and are hereinafter referred to as \textit{clients}. Learning in a federated setting involves several clients iteratively training local models on their data, which are then aggregated by a central server to form a single global model that is used by each client as an initialization for the next round of local training.

Given the sensitive nature of medical data, federated learning has naturally been utilized in a variety of healthcare applications \cite{digital}. For instance, it has been used for detecting COVID-19 lung abnormalities \cite{covid1} and for brain tumor segmentation \cite{bra2}. In these studies\cite{covid1}\cite{bra2}, federated learning demonstrated comparable performance to conventional data sharing approaches while mitigating privacy concerns. To the best of our knowledge, no federated learning approach has yet been applied to surgical videos, which is the application context of this work.

\subsection{Federated Learning with Unlabeled Data}

While federated learning bypasses the need to share data and facilitates collaborative training, the availability of labeled data still constitutes a major bottleneck, especially in healthcare applications. Participating in a supervised federated learning network requires each client to label their data. This is often prohibitively expensive in the medical domain, where data labeling commonly requires clinical expertise. Even assuming the availability of this expertise, the review and coordination on consistency across multiple clients are not always feasible without shared data. This is a major concern given the complexity and inherent ambiguities that several medical applications present.

Straightforward solutions may involve naive adaptations of generic semi-supervised methods \cite{sohn2020fixmatch,xie2019unsupervised,berthelot2019mixmatch} or those that were proposed for surgical phase recognition \cite{yengera2018less,dipietro2019automated,yu2018learning,shi2021semi}. However, these methods were primarily designed for clustered datasets, and hence such adaptations may not be suitable if the labeled data and unlabeled data come from different distributions. Still, the few federated learning works that address label deficiency \cite{jin2020towards,ji2021emerging,iqbal2021concepts}, demonstrate the feasibility and value of exploiting unlabeled data from multiple sources.

\subsubsection{\textcolor{blue}{Federated Unsupervised Representation Learning}}

Federated unsupervised data representation learning was studied in \cite{zhang2020federated,zhuang2021collaborative}, in which multiple clients collaborate to learn useful data representations on unlabeled data before leveraging this learned knowledge in supervised downstream tasks, such as classification. Correspondingly, federated unsupervised representation learning has also demonstrated value in medical imaging, specifically for detecting COVID-19 from x-ray chest scans \cite{dongnan}, cardiac MRI segmentation \cite{wu2021federated}, and brain MR anomaly segmentation \cite{bercea2021feddis}.

\subsubsection{\textcolor{blue}{Federated Unsupervised Domain Adaptation}}

Unsupervised domain adaptation has been also formulated in a federated setting \cite{peng2019federated, song2020privacy,yao2022federated} to preserve privacy, in which clients with labeled data collaborate with clients with unlabeled data to enhance training for the latter. 

\subsubsection{\textcolor{blue}{Federated Semi-Supervised Learning}}

A similar training scenario involving both labeled and unlabeled data, under which our works falls, is Federated Semi-Supervised Learning (FSSL). In FSSL, clients with labeled and/or unlabeled data collaborate to enhance training for all clients. FSSL has been addressed in several works, most of which utilize pseudo-labeling, a technique to automatically generate artificial labels using model predictions. \cite{albaseer2020exploiting} first introduced a simple framework utilizing pseudo-labeling in a federated setting. Other works proposed methods to enhance pseudo-labels. \cite{che2021fedtrinet} utilized a dynamic thresholding strategy for selecting pseudo-labels based on model confidence along with having multiple clients vote to generate pseudo-labels. \cite{bdair2021fedperl} utilized peer learning and ensemble averaging from multiple clients. The aforementioned works primarily addressed the scenario where clients have partially labeled datasets, i.e. all clients have both labeled and unlabeled data. In contrast, our work addresses the scenario where a client with a fully labeled dataset and multiple clients with fully unlabeled datasets collaborate for FSSL.

This formulation of labeled/unlabeled datasets decoupling has been only recently garnering attention \cite{jeong2020federated,zhang2021improving,long2020fedsiam,diao2021semifl,long2021fedcon,liu2021federated,yang2021federated}, but could have crucial practical implications on the scalability of models for healthcare applications by facilitating the participation of clients with completely unlabeled data. This scenario is referred to as FSSL with \textit{global semi-supervision} \cite{yang2021federated}. In other applications where large labeled datasets are more readily available or can be transferred, the roles of the aggregating server and the labeled data owner may overlap, and this scenario is instead referred to as the \textit{labels-at-server} scenario\cite{jeong2020federated ,long2020fedsiam,zhang2021improving,long2021fedcon,diao2021semifl}.

\cite{jeong2020federated} introduced an inter-client consistency loss for mitigating domain shifts and model decomposition for supervised-unsupervised learning decoupling. \cite{zhang2021improving} proposed a method to minimize gradient diversity across clients models by replacing batch normalization with group normalization \cite{wu2018group} and a new model averaging alternative to federated averaging. \cite{long2020fedsiam} proposed an adaptive layer-wise parameter selection method for uploading models for aggregation. Other works that also achieve competitive performance on benchmark datasets include \cite{long2021fedcon,liu2021federated,diao2021semifl}. \cite{diao2021semifl} adapted a combination of two state-of-the-art semi-supervised methods, FixMatch \cite{sohn2020fixmatch} and MixMatch \cite{berthelot2019mixmatch}, in a federated setting. \cite{long2021fedcon} and \cite{liu2021federated} proposed methods based on contrastive learning and knowledge distillation, respectively. Apart from classification, FSSL has been also used for the task of COVID-19 region segmentation in chest computed tomography scans \cite{yang2021federated}.

A crucial aspect of our work is the importance of learning relevant temporal information from surgical videos, an issue that has not explicitly been addressed in any of the above works. Temporal modeling in federated learning and, more specifically, in FSSL is still relatively unexplored. Only a few applications exist such as traffic flow forecasting \cite{meng2021cross}, human activity recognition \cite{zhao2020semi, presotto2021semi}, audio recognition \cite{tsouvalas2021federated}, and machine fault diagnosis \cite{zhang2021federated}.

Finally, it is worth noting that in all the FSSL mentioned methods except \cite{yang2021federated}, data distribution shifts across clients were simulated artificially based on label distributions. Aside from label distribution differences, our data are also characterized by a heterogeneity that originates from the different hospitals, such as surgical workflow, demographic, and hardware variability. This practical difference highlights the need for further study on real-world data, which has not been well investigated in previous works.

\section{FedCy: Directed cycle consistency  using federated semi-supervised learning}
\label{sec:methods}

In this section, we describe our proposed federated semi-supervised learning method for surgical phase recognition. Surgical phase recognition is a single-label multiclass classification problem of surgical video frames. Though distinct scenarios have been described in previous works\cite{bdair2021fedperl,jeong2020federated,yang2021federated} of federated learning using varying amounts of labeled data at different locations, our work specifically assumes the presence of one fully labeled private dataset and several other completely unlabeled private datasets. Whereas our proposed method can be extended to utilize multiple labeled datasets, we choose to tackle this instance of label deficiency due to the practical limitations of generating consistent annotations for complex tasks without shared data and review processes. This represents the real-world use-case allowing to scale models to clients that do not have the technical and clinical bandwidth to generate consistently labeled datasets.

As discussed in Section \ref{sec:context} and empirically proved in our results, previous FSSL approaches may be suboptimal for the task of phase recognition. We, therefore, design our method to learn temporal patterns found in videos while effectively harnessing the task knowledge that can be learned from the labeled data. We adapt Temporal Cycle Consistency \cite{dwibedi2019temporal}, a self-supervised learning technique for learning temporal patterns, by guiding it to learn more task-relevant features through concurrently optimizing a contrastive loss on the labeled data. In the rest of this section, we first describe two core components of FedCy - Temporal Cycle Consistency and constrastive learning - and finally the federated training process.

\subsection{Temporal Cycle Consistency: Unsupervised learning of temporal correspondences}
\label{sec:tcc}
Temporal Cycle Consistency Learning (TCC) \cite{dwibedi2019temporal} is a self-supervised method for learning temporal correspondences between videos. TCC has been shown to be useful as a self-supervised pretraining task for learning spatio-temporal representations that can boost performance  on downstream tasks such as activity recognition in videos\cite{dwibedi2019temporal}. In this subsection, we briefly describe the concept of cycle consistency learning, which we later adapt for the task of surgical phase recognition.

Consider two sequences of video frames: $S = \{s_{k}\}_{k=1}^{n}$ and $T = \{t_{k}\}_{k=1}^{m}$. Given a feature extractor $\phi$, let $U = \{u_k\}_{k=1}^{n}$ and $V = \{v_{k}\}_{k=1}^{m}$ denote the lower dimensional embeddings of the frames of $S$ and $T$ (i.e. $u_k = \phi(s_{k}), v_k = \phi(t_{k})$), respectively. Finally, let $\mathcal{Q}$ denote a similarity metric of two feature vectors (e.g. cosine similarity).\\

\subsubsection{Principle}
For an embedding $u_k \in U$, let:

\begin{itemize}
    \item $v_r$ denote the nearest neighbor of $u_k$ in $V$.
    \item $u_l$ denote the nearest neighbor of $v_r$ in $U$.
\end{itemize}
An embedding $u_k$ is said to be \textit{cycle consistent} if $l = k$. \cite{dwibedi2019temporal} reformulated this constraint as a regression task with a differentiable loss function, which we describe below, that can be used as a learning objective to train deep neural networks.\\

\subsubsection{Consistency loss for a pair of sequences (S,T)} For an embedding $u_k \in U$, we compute its soft nearest neighbor $\Tilde{v}$ in $V$:

\begin{equation}
    \label{eq:tcc2}
    \Tilde{v} = \sum_{i=1}^{m} \alpha_i v_i \hspace{8pt}\text{, where}\hspace{8pt}\alpha_i = \frac{e^{\mathcal{Q}(u_k, v_i)/\tau}}{\sum_{j=1}^{m} e^{\mathcal{Q}(u_k, v_j)/\tau}}.
\end{equation}

\noindent Here, we use the softmax function to compute $\alpha = \{\alpha_i\}_{i=1}^m$, a vector representing the similarity of $u_k$ to each of the embeddings of $V$, with $\tau$, the softmax temperature that is used to scale the logits fed into the softmax function. Similarly, we compute $\beta = \{\beta_i\}_{i=1}^n$, the vector representing the similarities of $\Tilde{v}$ to each of the embeddings of $U$. For $u_k$ to be cycle consistent with respect to V, $\beta$ would need to show peak-like behavior at its $k^{th}$ entry. The loss function described below enforces this using a Gaussian prior imposed on $\beta$:

\begin{equation}
    \label{eq:tcc4}
    \mathcal{L}(u_k, V) = \frac{(k-\mu)^2}{\sigma^2} + \lambda_{\sigma} \log(\sigma),
\end{equation}

\noindent where $\mu = \sum_{i=1}^ni\beta_i$, $\sigma=\sum_{i=1}^n(i-\mu)^2\beta_i$, and $\lambda_{\sigma}$ is the weight for the variance regularization loss term that forces $\beta$ to be more sharp around its $k^{th}$ entry. The final consistency loss for two sequences of frames $S,T$ is thus:

\begin{equation}
    \label{eq:tcc5}
    \mathcal{L}_{\text{TCC}}(S, T) = \frac{1}{n+m}\bigg(\sum_{k=1}^n \mathcal{L}(u_k, V) + \sum_{k=1}^m \mathcal{L}(v_k, U) \bigg).
\end{equation}

\subsubsection{Clip Sampling}
\label{sec:clipsampling}
A straightforward implementation of cycle consistency loss would involve training using multiple videos. However, this approach is not easily scalable, particularly when dealing with long videos such as surgical recordings, due to memory constraints. Short clips are sampled from videos and used for training instead.

We define a clip as a set of chosen frame IDs $c$ selected per sampling strategy. Let $V$ be a video of $n$ frames, whose IDs are defined by the set $\{1,2,3,\dots ,n\}$, and let $k$ be the size of the clip to be sampled. In \cite{dwibedi2019temporal}, two sampling strategies were used in the case of long videos, albeit for much shorter durations than the surgical videos used in this work:\\
\noindent
\textbf{1) Uniformly Strided Sampling with Offset}:
\begin{equation*}
    c = \{o, o+s, o+2s, \dots, o+(k-1)s\},
\end{equation*}
\noindent
where $s>0$ is the stride (fixed) and $o$ is a randomly (uniformly) chosen offset constrained by $o+(k-1)s \leq n$.\\
\noindent
\textbf{2) Random Sampling with Offset}:
\begin{equation*}
    c = \{s_1, s_2, \dots, s_{k}\},
\end{equation*}
\noindent
where $\{s_i\}_{i=1}^k$ are randomly (uniformly) chosen samples constrained by a fixed offset $o$: $s_i \geq o$ $\forall i$ and $s_i > s_j$ $\forall i > j$.

Laparoscopic cholecystectomy videos are usually characterized by highly variable phase durations across different videos \cite{twinanda2016endonet}. We hypothesize that using a fixed stride (sampling strategy \textbf{1}) thus may not be efficient in our case. Additionally, since some phases are much shorter than others, sampling strategy \textbf{2} may not efficiently represent entire procedures during training.

\begin{figure}[H]
    \centering
    \includegraphics[trim={0cm 0cm -1cm 0cm},clip,width=0.5\textwidth, keepaspectratio]{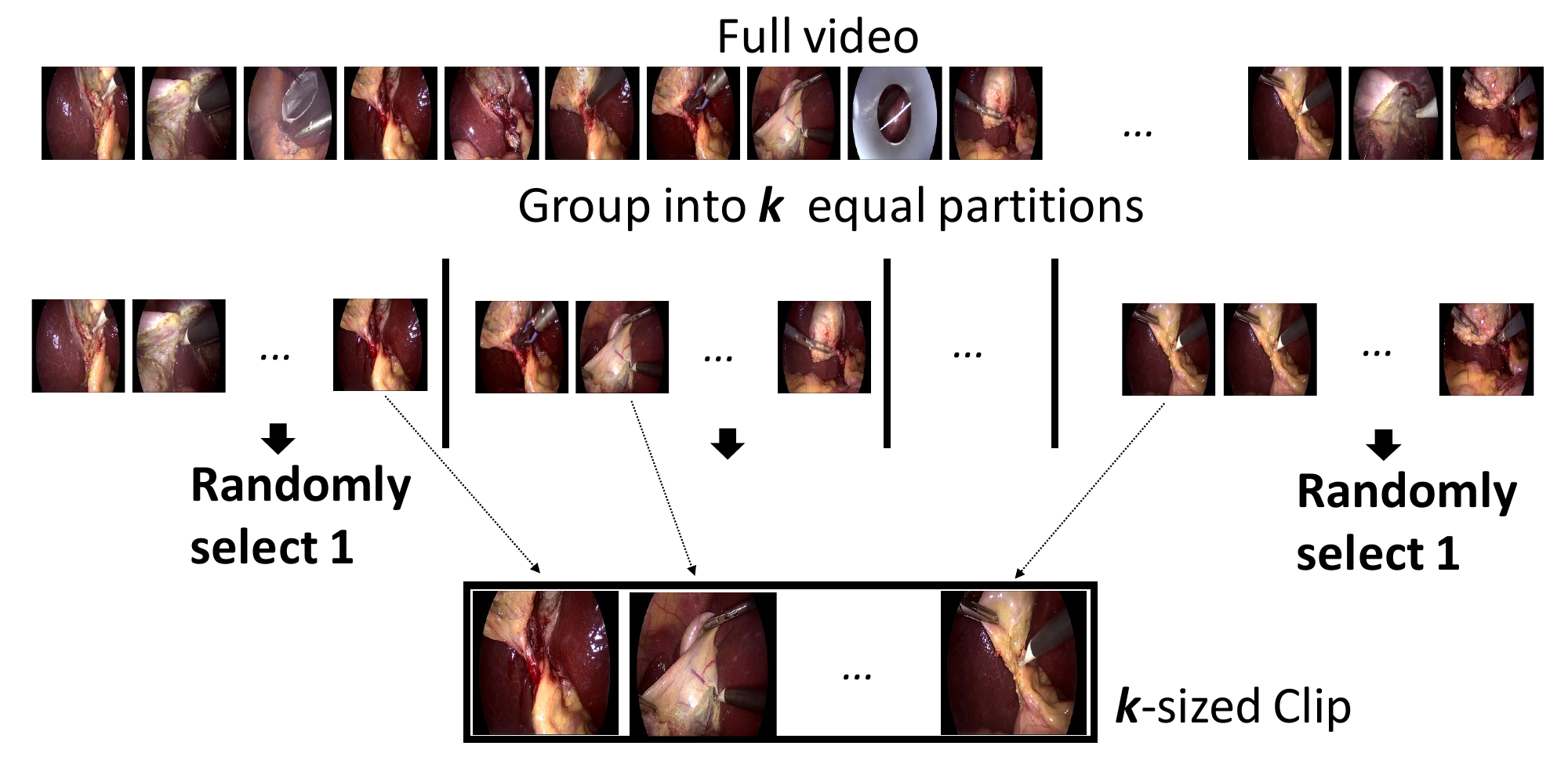}
    \caption{Our proposed clip sampling strategy.}
    \label{fig:clip}
\end{figure}

We therefore introduce a clip sampling strategy that mitigates the above concerns, depicted in Fig. \ref{fig:clip}. Each video is divided into $k$ equal partitions, $k$ being the clip size to be sampled. Then, $1$ frame is randomly selected from each partition to generate a clip of size $k$. We sample $\lfloor \frac{L}{k} \rfloor$ non-overlapping clips from each video to be used in each epoch of training, where $L$ is the corresponding video length.

\begin{figure*}[ht]
    \centering
    \includegraphics[trim={-0.7cm 5cm 0cm 0.9cm},clip,scale=0.50, keepaspectratio]{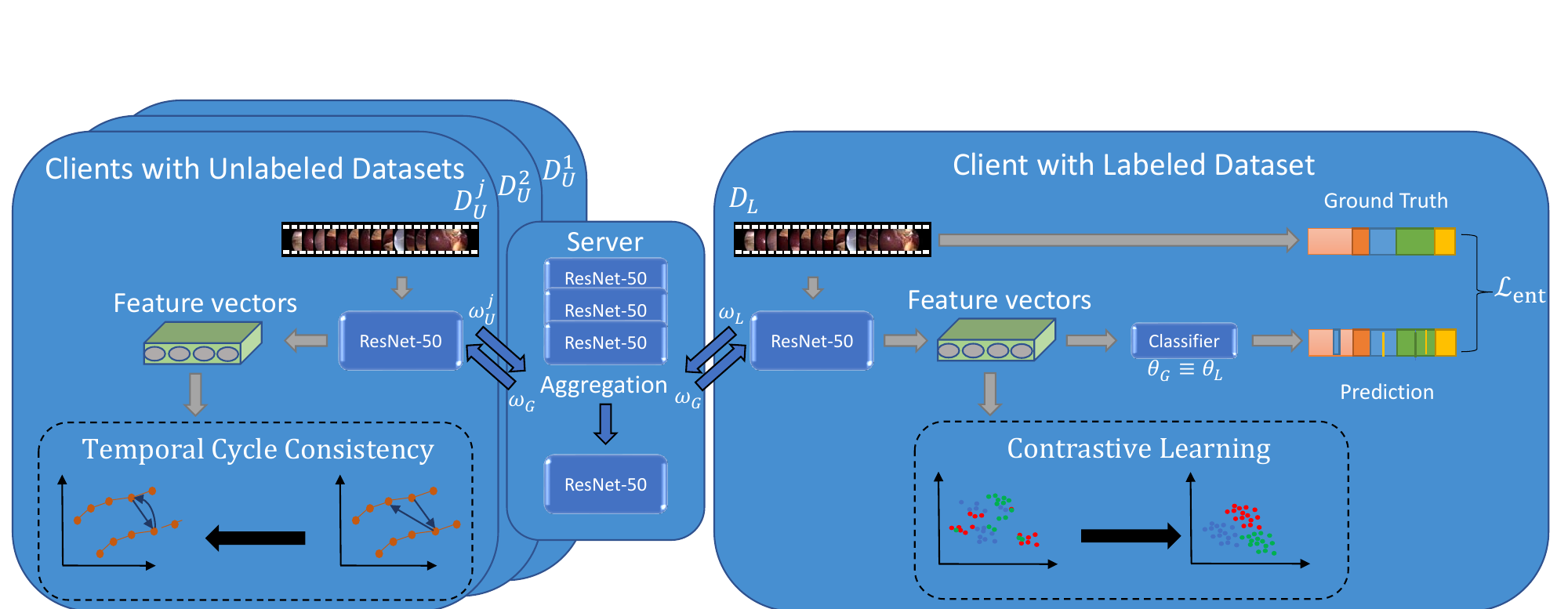}
    \caption{A brief overview of the FedCy training procedure. A feature extractor \textcolor{blue}{(ResNet-50)} is trained on each unlabeled dataset by minimizing the temporal cycle consistency loss, while a feature extractor and a classifier are trained on the labeled dataset by minimizing a contrastive loss and a cross-entropy loss. The feature extractors are aggregated by the central server after each federated learning round.}
    \label{fig:fedcy}
\end{figure*}

\subsection{Contrastive Learning}

Contrastive learning was first introduced in \cite{1640964}. Given a frame $d_a$ (the \textit{anchor}), a \textit{positive} frame $d_p$ (e.g. a frame that belongs to the same class of $d_a$, or an augmented version of $d_a$), and a \textit{negative} frame $d_n$, the principle is to learn a feature representation $\phi$ that maps $d_a, d_p, d_n$ to feature vectors $f_a, f_p, f_n$ while maximizing the similarity of $(f_a, f_p)$ and minimizing the similarity of $(f_a, f_n)$.

In this work, we will be using the NT-xent loss function for contrastive learning, as termed in \cite{chen2020simple}. For an \textit{anchor} $f_a$, a \textit{positive} $f_p$ and a set of \textit{negatives} $F_n$:

\begin{equation}
    \label{eq:cont}
    \mathcal{L}_{\text{NT}}(f_a, f_p, F_n) = -\log\bigg(\frac{e^{\mathcal{Q}(f_a, f_p)/\tau}}{\sum_{f \in F_n \cup \{f_p\}} e^{\mathcal{Q}(f_a, f)/\tau}}\bigg),
\end{equation}

\noindent where $\tau$ is the NT-xent temperature.

\subsection{Learning Objective}

\subsubsection{Problem formulation} Given a single client $C_L$ with a private local dataset $D_{L}$ and a set of M clients $C_U = \{C_U^j\}_{j=1}^M$ with private unlabeled datasets $\{D_U^j\}_{j=1}^M$, our aim is to learn a single global model that is effective on all the clients for phase recognition. The labeled dataset $D_{L}$ can be represented by: $D_{L} = \{(x_i, y_i)\}_{i=1}^n$, with $n$ denoting the total number of frames, $x_i$ denoting a frame, and $y_i$ a one-hot label corresponding to one of $P$ phases (classes) represented in $x_i$. Each unlabeled dataset, $D_U^j$, can be represented as a set of clips $ \{c_i\}_{i=1}^{n_c^j}$ where $c_i$ is a clip of $k$ frames and $n_c^j$ is the total number of clips in $D_U^j$ sampled according to the strategy defined in Section \ref{sec:clipsampling}. All the considered clients collaboratively train, in a federated setting coordinated by a central server, a global model represented by a feature extractor $\phi_{G}$ parameterized by $\omega_G$ and a classifier $\mathcal{H}_G$ parameterized by $\theta_G$, that produces softmax outputs corresponding to phase probabilities.

The clients $C_U$ will learn spatio-temporal information which, when supported by the supervised discriminative knowledge learned by the client $C_L$, can be useful for phase classification without the need for client-specific fine-tuning. As illustrated by Fig. \ref{fig:fedcy}, the training procedure is divided into three main steps detailed below and executed repeatedly as follows:

\begin{outline}
 \1 In Parallel:
   \2 \textbf{Unsupervised training} by each client $C_U^j$.
   \2 \textbf{Supervised training} by the client $C_L$.
 \1 \textbf{Model aggregation} by the central server.
\end{outline}

\subsubsection{Unsupervised Training}
Each client $C_U^j$ locally trains a feature extractor $\phi_U^j$ parameterized by $\omega_U^j$ by minimizing the temporal cycle consistency loss on the clips of its dataset using equations \ref{eq:tcc2}--\ref{eq:tcc5}. For a given batch of clips $\mathcal{B} \subset D_U^j$, each client optimizes the following:

\begin{equation}
    \min_{\omega_U^j} \frac{1}{|\mathcal{B}|(|\mathcal{B}|-1)} \sum_{\substack{c_i,c_k \in \mathcal{B} \\ i\neq k}} \lambda_t\mathcal{L}_{\text{TCC}}(c_i, c_k),
\end{equation}

\noindent where $\lambda_t$ is a weight.\\

\subsubsection{Supervised Training}
Due to the complexity of the phase recognition task, unsupervised learning of temporal correspondences may not necessarily yield spatio-temporal representations that are directly useful for phase recognition. For example, recurring events such as unexpected bleeding that are not specific to any phase of the procedure may serve as confounding factors during training. The objective function of the client $C_L$ is thus designed in a way that it can play the role of a guide for the unsupervised training being done by the other clients. To do this, in addition to minimizing the supervised cross-entropy loss, we force representations of different classes to be more distant in the feature space by adding a contrastive loss term on the final objective of this client. \textcolor{blue}{We hypothesize that the parallel training objectives during each training round - the contrastive loss for supervised training and the TCC loss for unsupervised training - will complement each other. }Further, we believe that propagating this learned knowledge through federated learning will drive the network towards finding more task-specific temporal correspondences from the unlabeled data.

Therefore, the client $C_L$ will train a feature extractor $\phi_{L}$ of parameters $\omega_{L}$ and a classifier $\mathcal{H}_{L}$ of parameters $\theta_{L}$ by minimizing a contrastive loss term and a cross-entropy loss term. A batch of frames $\mathcal{B} \subset D_{L}$ can be represented as $\mathcal{B} = \{J_1, J_2, \dots, J_P\}$ where $J_i$ contains the frames belonging to class $i$. Let $\{X_1, X_2, \dots, X_P\}$ denote the sets of the corresponding extracted feature vectors of $\mathcal{B}$ by $\phi_L$. Then, the contrastive loss on $\mathcal{B}$ can be calculated using equation \ref{eq:cont} as follows:

\begin{equation}
    \mathcal{L}_{\text{cont}}(\mathcal{B}) = \frac{1}{|\mathcal{B}|}\sum_{i=1}^P \mathcal{L}(X_i),
\end{equation}

\noindent where:

\begin{equation}
    \mathcal{L}(X_i) = \sum_{f_a \in X_i} \frac{1}{|X_i|-1} \sum_{\substack{f_p \in X_i \\ f_p \neq f_a}} \mathcal{L}_{\text{NT}}(f_a, f_p, \bigcup_{j \neq i} X_j).
\end{equation}

\noindent The final objective of the client $C_L$ will thus be:

\begin{equation}
    \min_{\omega_{L},\theta_{L}} \frac{1}{|\mathcal{B}|} \sum_{(x,y) \in \mathcal{B}} \mathcal{L}_{\text{ent}}(y, \mathcal{H}_{L}(\phi_{L}(x))) + \lambda_{c}\mathcal{L}_{\text{cont}}(\mathcal{B}),
\end{equation}

\noindent where $\mathcal{L}_{\text{ent}}$ is the cross-entropy loss function, and $\lambda_c$ is the weight of the contrastive loss term.\\

\subsubsection{Model Aggregation at the Server}
After each round of local training, the server will aggregate the trained local models using FedAvg \cite{pmlr-v54-mcmahan17a} into a global model $(\phi_G,\mathcal{H}_G)$ which will then be sent back for another local training round:

\begin{equation}
    \omega_G = p_L\omega_L + \sum_{j=1}^M p_U^j\omega_U^j \hspace{8pt}\text{and}\hspace{8pt} \theta_G = \theta_L,
\end{equation}

\noindent where $p_L$ and $p_U^j$ represent respectively the fraction of data contributed by clients $C_L$ and $C_U^j$ during each round. \textcolor{blue}{We note that during training, classifier weights are used only by the client holding labeled data. During deployment, the classifier weights will be simply shared across all the clients.}\\

\section{Datasets}

\subsection{MultiChole2022: A Multicenter Laparoscopic Cholecystectomy Dataset}

To evaluate FedCy, we introduce Multicenter Cholecystectomy 2022 (MultiChole2022): a large multicenter dataset comprising 180  laparoscopic cholecystectomy (LC) videos. MultiChole2022 is composed of the 80 videos of the public dataset Cholec80 \cite{twinanda2016endonet}, which were collected from the University Hospital of Strasbourg, France, along with 4 sets of 25 videos collected in the following Italian hospitals: Policlinico Universitario Agostino Gemelli, Rome; Azienda Ospedaliero-Universitaria Sant’Andrea, Rome; Fondazione IRCCS Ca' Granda Ospedale Maggiore Policlinico, Milan; and Monaldi Hospital, Naples \textcolor{blue}{ through the AI4SafeChole Consortium.} Participating hospitals only shared anonymized endoscopic videos through encrypted hard drives. No clinical data were harvested or shared. Datasets of these hospitals will be anonymously denoted by $D_{1-4}$.

The inherent data diversity represented in MultiChole2022 could facilitate research in a variety of topics such as domain adaptation and federated learning for laparoscopic surgery. Table \ref{tab:stat1} presents the total, mean, and standard deviation of the durations of the videos of each hospital.
\begin{table}[H]
  \begin{center}
    \begin{tabular}{cccc}
    \hline
      \textbf{Dataset} & \textbf{Total} & \textbf{Mean} & \textbf{Std Dev}\\
      \hline
        Cholec80 & 51:09:39 & 00:38:22 & 00:16:59\\\hline
        $D_1$ & 17:45:30 & 00:42:37 & 00:18:39 \\\hline
        $D_2$ & 16:27:07 & 00:39:29 & 00:25:14 \\\hline
        $D_3$  & 18:11:58 & 00:43:41 & 00:12:21 \\\hline
        $D_4$ & 14:22:37 & 00:34:30 & 00:10:27 \\\hline

    \end{tabular}
        \caption{Duration Statistics (in HH:MM:SS) of videos of each participating hospital.}
    \label{tab:stat1}
  \end{center}
\end{table}

\subsection{Data Annotation}

LC workflow has been divided into phases in several ways \cite{garrow2021machine}. In \cite{twinanda2016endonet}, a LC is divided into 7 phases: \textit{Preparation}, \textit{Calot Triangle Dissection}, \textit{Clipping and Cutting}, \textit{Gallbladder Dissection}, \textit{Gallbladder Packaging}, \textit{Cleaning and Coagulation}, and \textit{Gallbladder Extraction}.

To annotate the workflow in LC videos, a robust  protocol describing the visual clues signaling the boundaries of each phase should be defined to ensure reproducibility and consistency of annotations across annotators and hospitals \cite{mascagni2021artificial}.

\subsubsection{The MultiChole2022 phase annotation protocol} 
\begin{figure}[H]
    \centering
    \includegraphics[width=0.48\textwidth, keepaspectratio]{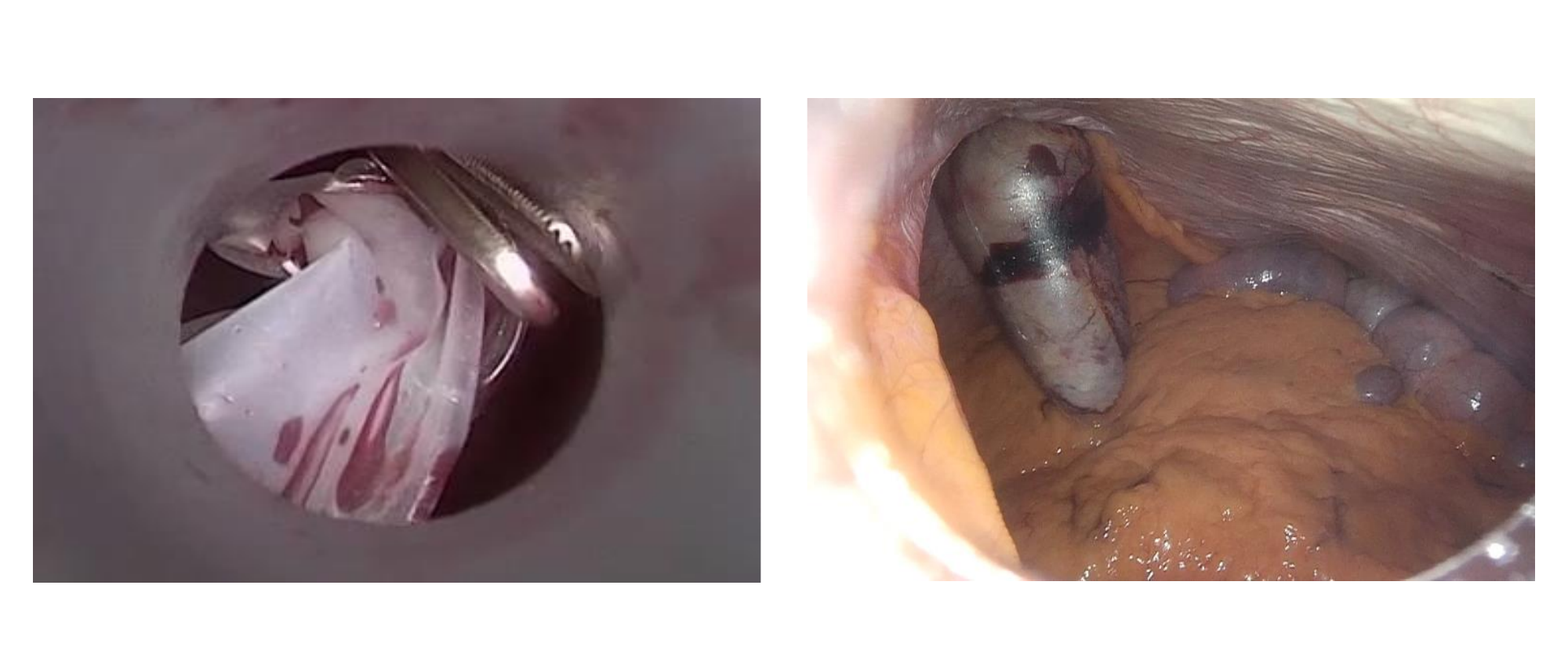}
    \caption{(Left) The grasper brings the specimen bag towards the camera trocar, through which the gallbladder will be extracted. (Right) The gallbladder is being pulled though a trocar other than the camera's.}
    \label{fig:hook}
\end{figure}

Surgical workflows may differ across hospitals, and hence annotation protocols designed for videos from a specific hospital might not generalize well to videos from another center. For instance, the protocol used to annotate Cholec80 made large use of instrument presence to define the boundaries of a phase, assuming a consistent use of instruments and sequence of phases across procedures. However, this is more likely to vary across hospitals. In addition, some phases might be performed differently across hospitals. Fig. \ref{fig:hook} illustrates the case of \textit{Gallbladder Extraction}. In some hospitals, the gallbladder is extracted through the same trocar used to insert the endoscopic camera whereas in others a lateral trocar is used. Consequently, in the first case, the gallbladder extraction phase largely takes place out-of-sight whereas in the second strong visual cues can be defined that consistently mark the beginning of this phase. Given these considerations, we designed a new protocol in order to annotate the diverse 180 MultiChole22 videos. In brief, the primary differences between the MultiCholec22 protocol and the protocol used in \cite{twinanda2016endonet} are: (1) Procedures are segmented into 6 phases rather than 7, excluding the Gallbladder Extraction phase for the aforementioned reasons; (2) Starting signals that mark the beginning of each phase are defined to focus more on actions rather than instrument usage to account for workflow variability.
\subsection{External Evaluation}
\label{sec:external data}
To evaluate how well FedCy performs on data from hospitals not participating in training, we also annotate a subset of 6 videos from the TUM LapChole dataset \cite{stauder2016tum} following the previously introduced annotation protocol. These 6 videos were taken from the Hospital Klinikum Rechts der Isar of Munich, Germany. \textcolor{blue}{The mean duration of these videos is 00:34:54, with a total of 03:29:23 and a standard deviation of 00:11:51.}

\section{Experimental Setup}
\subsection{Baselines}
\label{sec:baselines}
We establish several baselines to contextualize FedCy with respect to various learning paradigms. In all cases, our aim is to train a ResNet-50 \cite{he2016deep} model initialized from ImageNet \cite{russakovsky2015imagenet} pretrained weights for the task of surgical phase recognition. In all semi-supervised experiments, we use Cholec80\cite{twinanda2016endonet} as the labeled dataset and refer to the 4 unlabeled datasets as $D_{1-4}$.

Our first category of baselines comprises various fully- and semi- supervised methods assuming access to a single, fully labeled dataset and 4 completely unlabeled datasets, corresponding to 5 independent clients. These methods were chosen to demonstrate the superiority of our approach to both na\"ive federated approaches and state-of-the-art designs that use the same amount of labeled data. The first baseline, FullSup-Cholec80, is trained on a fully labeled cholec80 train set. This presents the standard non-federated approach that excludes external unlabeled data from training. The next baselines, FedFixMatch, FedUDA, and FedTCC represent na\"ive adaptations of FixMatch \cite{sohn2020fixmatch}, UDA \cite{xie2019unsupervised} and TCC pretraining \cite{dwibedi2019temporal}, respectively, using federated averaging \cite{pmlr-v54-mcmahan17a}. Our two final baselines in this category, FedMatch \cite{jeong2020federated} and FedRGD \cite{zhang2021improving}, are state-of-the-art FSSL approaches. Here, FedRGD proposes several changes to improve model performance including a change to the network design, replacing batch normalization layers with group normalization \textcolor{blue}{(GN)} \cite{wu2018group}. To perform a fair comparison with FedRGD, we also implement our final FedCy model with group normalization, denoted by FedCy-GN.

Our second category of baselines includes only fully supervised approaches assuming that each of the included datasets is fully labeled. These baselines, which are dependent on the presence of additional labeled data, are included to provide the target performance that we would like to achieve through this and future semi-supervised work. In this category, FullSup-Each, is a model trained using full supervision only on the dataset it is being evaluated on. This represents the strictest scenario in terms of privacy where no sharing of either data or models is involved in training. In contrast, FullSup-All, represents the most relaxed scenario, where a single model is trained on all the data collected together and treated as a single, centralized dataset. Finally, federated averaging \cite{pmlr-v54-mcmahan17a} represents the middle ground where models but not data can be transferred freely. 
\begin{table*}[ht]
     \centering
     \begin{tabular}{cccccccc}
        \hline
          & Cholec80 & $D_1$ & $D_2$& $D_3$ & $D_4$& Overall$_{\text{unlabeled}}$ & Overall$_{\text{all}}$ \\\hline
          FullSup-Cholec80 & $69.35 \pm 1.19$ & $48.99 \pm 1.03$ & $42.50 \pm 2.01$ & $50.91 \pm 2.45$ & $48.87 \pm 4.09$ & $47.82 \pm 1.47$ & $52.12 \pm 1.41$\\\hline
          FedFixMatch & $68.26 \pm 0.98$ & $49.15 \pm 1.59$ & $41.29 \pm 3.12$ & $51.09 \pm 2.15$ & $47.77 \pm 1.61$ & $47.33 \pm 1.76$ & $51.51 \pm 1.60$\\
          FedUDA & $70.57 \pm 0.15$ & $48.70 \pm 1.05$ & $42.60 \pm 2.85$ & $50.11 \pm 0.75$ & $49.82 \pm 2.38$ & $47.81 \pm 1.21$ & $52.36 \pm 0.99$\\
          FedTCC & \textcolor{blue}{\pmb{$71.54 \pm 0.81$}} & \textcolor{blue}{$55.46 \pm 2.21$} & \textcolor{blue}{$46.29 \pm 4.20$} & \textcolor{blue}{$55.48 \pm 1.86$} & \textcolor{blue}{$53.26 \pm 1.78$} & \textcolor{blue}{$52.62 \pm 0.36$} & \textcolor{blue}{$56.41 \pm 0.37$}\\
          FedMatch & $63.77 \pm 0.41$ & $50.38 \pm 0.99$ & $42.60 \pm 1.86$ & $47.86 \pm 0.41$ & $50.27 \pm 0.83$ & $47.78 \pm 0.99$ & $50.98 \pm 0.73$\\
          FedRGD & $70.42 \pm 0.37$ & $55.04 \pm 1.99$ & $49.71 \pm 2.73$ & $54.17 \pm 1.17$ & $50.77 \pm 1.30$ & $52.42 \pm 1.71$ & $56.02 \pm 1.36$\\
          FedCy (Ours)    & $70.97 \pm 0.24$ & $58.34 \pm 1.28$ & $53.05 \pm 0.70$ & \pmb{$59.97 \pm 1.19$} & \pmb{$56.12 \pm 1.57$} & $56.87 \pm 0.90$ & \pmb{$59.69 \pm 0.75$}\\
          FedCy-GN (Ours) & $68.76 \pm 1.14$ & \pmb{$59.52 \pm 0.90$} & \pmb{$57.57 \pm 2.49$} & $56.30 \pm 0.59$ & $54.14 \pm 0.86$ & \pmb{$56.89 \pm 0.80$} & $59.26 \pm 0.86$\\\hline
        FullSup-Each & $69.35 \pm 1.19$ & $63.30 \pm 1.54$ & $56.16 \pm 1.52$ & $58.51 \pm 0.38$ & $69.68 \pm 1.89$ & $61.91$ & $63.40$ \\
          FedAvg & $73.96 \pm 0.62$ & $66.81 \pm 0.42$ & $58.86 \pm 1.89$ & $64.30 \pm 0.91$ & $64.91 \pm 1.19$ & $63.72 \pm 0.96$ & $65.77 \pm 0.89$\\
          FullSup-All & $72.48 \pm 1.82$ & $70.69 \pm 0.51$ & $63.34 \pm 2.96$ & $67.21 \pm 0.97$ & $72.70 \pm 0.43$ & $68.49 \pm 0.77$ & $69.29 \pm 0.97$\\\hline
     \end{tabular}
     \caption{Comparison of our proposed method FedCy to various baselines and state-of-the-art designs. We present the F1-score performance on each of the 5 datasets, averaged across the unlabeled datasets, and averaged across all considered datasets. From top to bottom, the horizontal lines separate the fully-supervised method using only Cholec80 data, the semi-supervised methods using labeled cholec80 data + unlabeled data from other clients, and the fully supervised methods using labeled data from all clients. We report mean and standard deviation across 3 runs.}
     \label{tab:r_baseline}
 \end{table*}
\subsection{Implementation details}
\label{details}
All presented models were trained on NVIDIA A100 GPUs. Videos were subsampled at 1 frame per second \textcolor{blue}{before doing the experiments to reduce the data size}. Soft data augmentation (shift, rotate, scale) was used in all experiments following \cite{czempiel2020tecno}. Each experiment, except the pretraining phase of FedTCC, was run for a minimum of 6 epochs and stopped when the validation F1-score stops improving for 3 consecutive epochs. FedTCC was pretrained for 30 epochs. The Cholec80 dataset was divided into splits containing 40-8-32 (training-validation-testing) videos following state-of-the-art usage. Each of the other 4 datasets in MultiChole2022 was divided into splits of 13-6-6 (training-validation-testing) by stratified random sampling. In all semi-supervised experiments, only the Cholec80 validation split was used. The validation splits of all the datasets were only used for the fully supervised baselines.

We use a fixed learning rate of $5\times 10^{-5}$ and a weight decay of $5\times 10^{-5}$ for all models based empirically on the FullSup-Cholec80 experiment. Similarly, we fix a batch size of $64$ for all models. Specifically, for FedCy and FedTCC, where unlabeled training happens on clips and not images, we set the clip size and batch size to 16 and 2, respectively. We use an Adam optimizer for all experiments, and a ResNet-50 \cite{he2016deep} feature extractor pretrained on ImageNet \cite{russakovsky2015imagenet} with a 2048-sized feature vector output. The classifier used on top of the ResNet-50 is a fully connected layer with a softmax activation. In FedRGD and FedCy-GN, batch normalization layers of the ResNet-50 are replaced with group normalization \cite{wu2018group} using $32$ channel groups. In these two models, group normalization layers weights are initialized with those of the batch normalization layers from the pretrained ImageNet weights. Finally, we also empirically set the TCC softmax temperature, NT-xent loss temperature, $\lambda_{c}$, and $\lambda_{t}$ to 0.05, 0.1, 10, and 10, respectively, based on validation results. In all federated experiments, the communication cost is set to one epoch of local training per round, and each client used a local optimizer whose weights are excluded from model aggregation.

\section{Experiments and Results}
\label{sec:results}

F1-score is used to evaluate all models. We present the test results per-client (Cholec80, $D_{1-4}$), macro-averaged across all unlabeled datasets (Overall$_{\text{unlabeled}}$), and macro-averaged across all datasets (Overall$_{\text{all}}$). Each of the presented results shows the mean and standard deviation of the model performance over 3 reruns \footnote{\textcolor{blue}{Note that FullSup-Each corresponds to several distinct models - one per center - and so, standard deviation is only presented per center.}}.

\subsection{Comparison to Baseline and State-of-the-Art Methods}

In Table \ref{tab:r_baseline}, we compare FedCy to several baselines, using varying amounts of supervision, as described in Section \ref{sec:baselines}. We firstly observe from FullSup-Cholec80, which is trained only on Cholec80 in a standard fully-supervised setting, that there are significant disparities between performance on different client datasets. In particular, the 21.5\% gap in F1-score when testing on videos from the same client versus the average performance on videos from the other 4 clients highlights a major limitation of the generalizability of deep learning models. The development of principled, systematic, and accessible approaches to both train and evaluate on data sourced by varied and independent clients represents a significant step towards the clinical translation of such applications. 

\textcolor{blue}{Federated learning-based approaches that can leverage the availability of unlabeled data from the other 4 clients expectedly show boosts in performance. For example, FedTCC which uses a ResNet-50 backbone that has been pretrained on all the unlabeled datasets using a federated averaging based implementation of TCC demonstrates an average boost of 4.8\% F1 on the unlabeled datasets over FullSup-Cholec80. In fact, among the 3 proposed methods that use federated learning adaptations of popular self-supervised learning methods (FedFixMatch, FedUDA, FedTCC), we identify that TCC is particularly suited to this task, with FedTCC outperforming more sophisticated approaches like FedMatch and FedRGD on the unlabeled datasets.} Our proposed approach, FedCy demonstrates markedly and consistently superior performance over all the presented FSSL approaches and the model trained on only Cholec80. Notably, this equates to an average increase of 4.2\% and 9\% F1 on the unlabeled datasets over the next best baseline in this category (FedTCC) and FullSup-Cholec80, respectively. We also see approximately the same overall performance of our approach whether it uses batch normalization or group normalization. Interestingly, despite using a large fully labeled dataset, FedCy also helps boost performance on the Cholec80 test set with an increment of 3.9\% F1 over fully supervised training on only Cholec80. This may be attributed to a need for even more labeled samples or a regularizing effect provided by the training procedure on the unlabeled data.

When comparing against the models trained using labeled data from every client, we see that FedCy goes a long way towards bridging the gap in performance towards fully supervised approaches while significantly mitigating privacy and annotations concerns. To note, the non-collaborative approach FullSup-Each, where each client trains a tailored model on their dataset, performs 5.9\% worse on average than training a global model trained on a single clustered dataset (FullSup-All). While this reflects the need for large and diverse datasets, this number may in fact understate this need because this baseline is predicated on the generation of independently but consistently annotated data. Impressively, on two of the five considered datasets, Cholec80 and D$_2$, FedCy even outperforms the FullSup-Each baseline by up to 1.6\%.

\subsection{Influence of TCC Parameters}

Given that FedCy demonstrates the significant value of semi-supervised learning of temporal correspondences, we carefully study \textcolor{blue}{the effect of parameter variations on} our formulation of TCC on unlabeled datasets. 

\subsubsection{Role of Clip size and batch size}

\begin{figure}[H]
    \centering
    \includegraphics[width=0.48\textwidth, keepaspectratio]{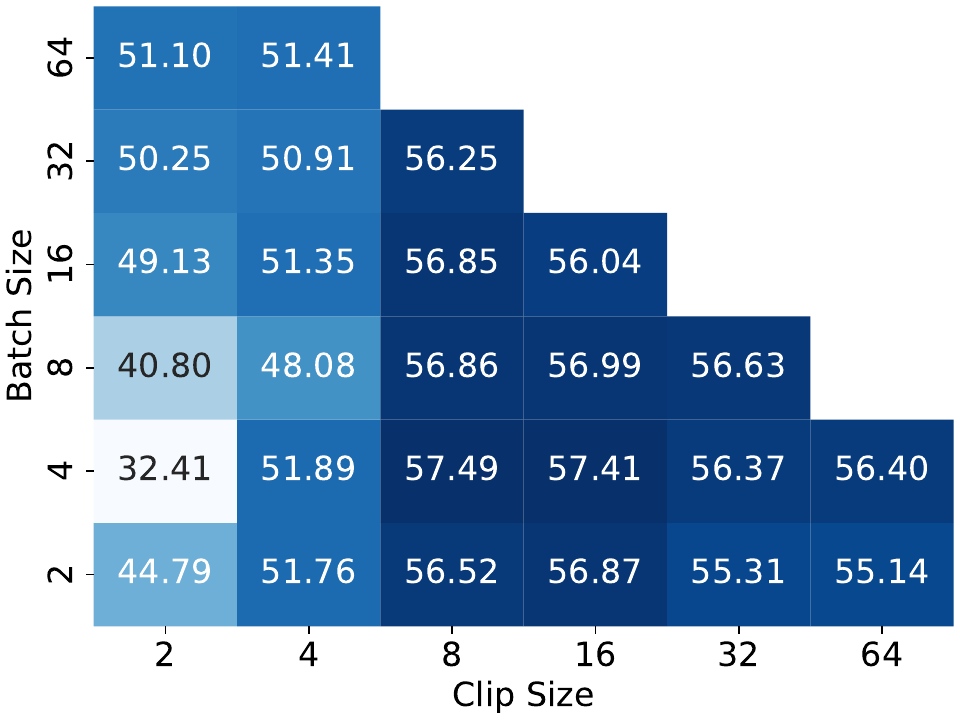}
    \caption{Overall F1-score performance on the unlabeled datasets for various clip and batch sizes. We report mean results across 3 runs.}
    \label{fig:r_clip}
\end{figure}

In this experiment, we analyze the role of clip size and batch size. Here, clip size reflects the granularity of temporal correspondences that we are trying to learn on the unlabeled data and batch size represents the number of clips we are trying to find correspondences between at each training iteration. In the heatmap presented in Fig. \ref{fig:r_clip}, we see large gains in performance up to a clip size of 8 after which the performance tends to saturate for all clip sizes. Concerning the batch size, we see that FedCy is largely robust to variations in this parameter.

\textcolor{blue}{We also mention here that GPU memory consumption per client during training varied approximately from 1GB to more than 60GB when moving diagonally in Fig. \ref{fig:r_clip} from (batch size $\times$ clip size = 4) to (batch size $\times$ clip size = 256), with 16GB being the value for our chosen parameters (batch size = 2, clip size = 16).}

\subsubsection{Sampling Strategies}
Here, we compare our proposed sampling strategy against the approaches used in \cite{dwibedi2019temporal} and described in Section \ref{sec:methods}. In Table \ref{tab:n_clip}, we see that the choice of sampling strategy could greatly affect results. Particularly, simply extending the strategies proposed in \cite{dwibedi2019temporal} to sample multiple clips from each video results in marked increase in performance. Our proposed sampling approach is $\sim 1$\% better, on average, than the next best strategy (random sampling using multiple clips).
\begin{table}[H]
     \centering
     \begin{tabular}{cccc}
        \hline
            & Ours & Uniformly Strided & Random \\\hline
           Single Clip & $50.82 \pm 1.22$ & $51.33 \pm 1.58$ & $50.08 \pm 1.76$\\\hline
           Multiple Clips & \pmb{$56.87 \pm 0.90$} & $52.09 \pm 1.16$ & $55.95 \pm 0.52$\\\hline

     \end{tabular}
     \caption{Effect of different sampling strategies on F1-score performance averaged across the unlabeled datasets. We report mean and standard deviation across 3 runs.}
     \label{tab:n_clip}
 \end{table}
 
\subsection{Ablation Study}

\textcolor{blue}{In this ablation study, we aim to prove the effectiveness of FedCy, both in terms of formulation and configuration. To this end, we quantify the performance improvement brought about by learning from unlabeled data using TCC with different configurations and then demonstrate how TCC can be utilized for more effective representation learning using the contrastive learning-based approach presented in FedCy. In Fig. \ref{fig:ablation}, we vary the configuration of TCC presented in our formulation of FedCy using the following baselines: (1) Centralized pretraining of the ResNet-50 using TCC for each unlabeled dataset before finetuning on Cholec80 (Vanilla TCC) (2) Pretraining simultaneously on all the unlabeled datasets using a federated averaging based implementation of the TCC loss before finetuning on Cholec80 (FedTCC) (3) Training end-to-end using federated averaging with a cross entropy loss on Cholec80 and the TCC loss on the unlabeled datasets (FedCy). For all these experiments, we use our proposed sampling strategy for clip sampling and the default hyperparameters described in \ref{details}. We also present each baseline with and without a contrastive loss applied simultaneously to the supervised cross-entropy loss on the labeled dataset.}

\subsubsection{Contribution of contrastive training objective}

\textcolor{blue}{In this subsection, we aim to illustrate the crucial role the supervised contrastive loss plays in boosting performance. Fig. \ref{fig:ablation} shows a clear improvement in the F1-score performance of FedTCC and FedCy experiments with the addition of the contrastive loss. Notably, we also see that even without the contrastive loss, FedCy and FedTCC perform on par with the state-of-the-art methods for federated semi-supervised learning.}

\subsubsection{Interplay between training objectives}

\begin{figure}[H]
    \centering
    \includegraphics[width=0.48\textwidth, keepaspectratio]{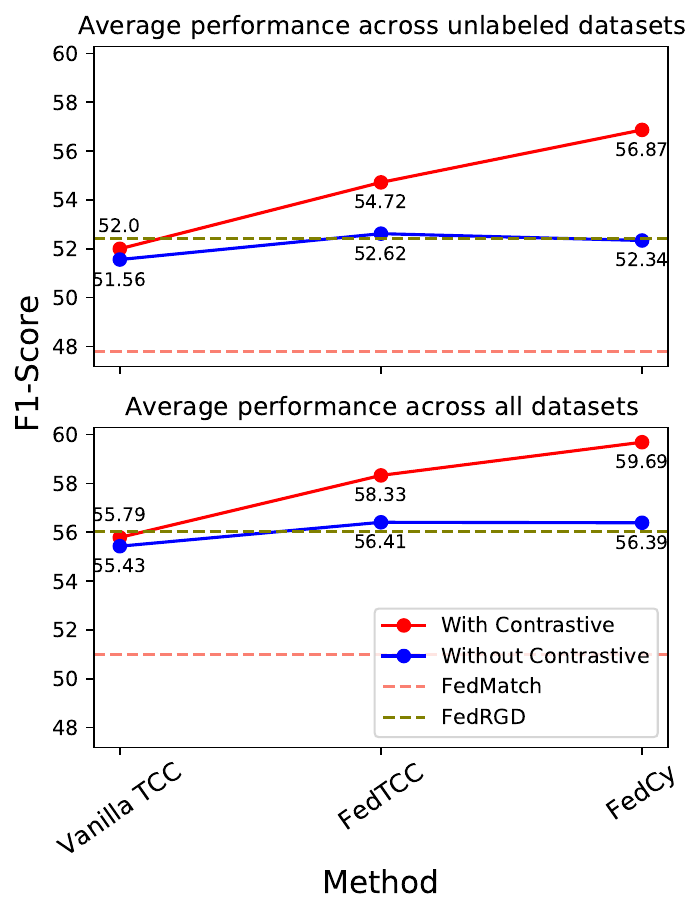}
    \caption{\textcolor{blue}{A comparison of FedCy to different configurations of TCC and contrastive learning. From left to right, per-center TCC pretraining (Vanilla TCC), federated averaging-based TCC pretraining (FedTCC), and FedCy. We report mean results across 3 runs.}}
    \label{fig:ablation}
\end{figure}

\textcolor{blue}{In this subsection, we aim to demonstrate the role of contrastive learning in guiding the unsupervised TCC training process in our proposed configuration of FedCy. By looking at Fig. \ref{fig:ablation}, we make the following points:}

\begin{itemize}
    \item \textcolor{blue}{The performance among the three configurations without the contrastive loss (blue plot) is comparable (within $\sim1\%$ F1), while those with the contrastive loss (red plot) show largely varying magnitudes of performance boosts ($\sim4-5\%$ F1). This demonstrates that the chosen configuration of contrastive loss and TCC can drastically influence results.}
    \item \textcolor{blue}{ Using the contrastive loss as a means to parallelly guide the training on the unlabeled data toward identifying ``task-informed'' temporal correspondences (FedCy) critically provides the largest boosts in performance. This corresponds to an average boost of $4.5\%$ and $3.3\%$ F1-score over the unlabeled datasets and all datasets, respectively.}
\end{itemize}

\subsection{Effect of Varying Labeled Data Size}
\begin{figure}[H]
    \centering
    \includegraphics[width=0.48\textwidth, keepaspectratio]{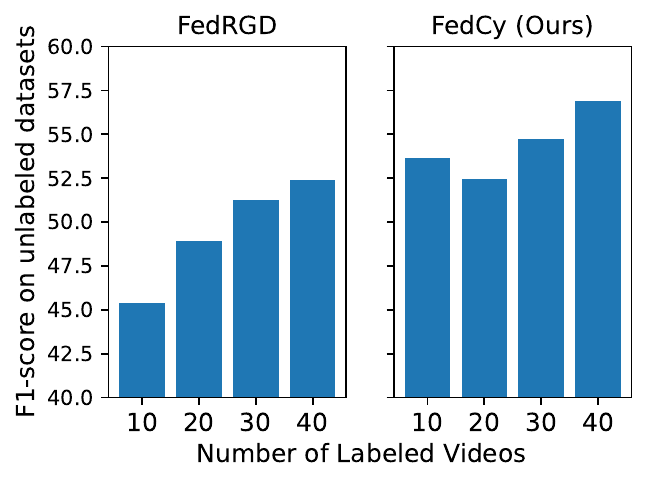}
    \caption{\textcolor{blue}{Overall F1-score performance on unlabeled datasets of FedCy and FedRGD with different labeled data size. We report mean results across 3 runs.}}
    \label{fig:less}
\end{figure}

\textcolor{blue}{To investigate the robustness of FedCy to labeled data deficiency, we varied the size of the training split of the labeled dataset (Cholec80). The original split (40 videos) was divided into 4 equal subsets using stratified random sampling. No change was made to the validation and testing splits. In Fig. \ref{fig:less}, we present FedCy and the best state of the art (FedRGD) with different labeled training data size. Note that the experiments with 40 videos are those presented in Table \ref{tab:r_baseline}. We report the overall performance on the unlabeled datasets.}

\textcolor{blue}{Interestingly, FedCy outperforms the top performing state-of-the-art method (FedRGD) at all considered amounts of labeled data. In particular, we observe that FedCy is able to better leverage lower amounts of labeled data to significantly boost performance on the unlabeled datasets. Going from 40 to 10 labeled videos, we observe a 7\% drop of FedRGD performance on unlabeled datasets, we see that FedCy is significantly more robust with only a 3.3\% drop in F1-score. In fact, using as few as 10 labeled videos, FedCy slightly outperforms FedRGD with the all 40 videos.
}

\subsection{External Evaluation}

\textcolor{blue}{In Table \ref{tab:external}, we see that both formulations of FedCy, with and without group normalization, outperform state-of-the-art methods for federated semi-supervised learning and are promisingly more robust to the held-out (out-of-distribution) dataset presented in Section \ref{sec:external data}. To ensure a fair comparison, we also reran all the baselines in Table \ref{tab:external} with group normalization, and note that FedCy-GN still outperforms the next highest semi-supervised benchmark (FedTCC + group normalization) by  $1.8\%$.  Interestingly, we see wide performance variations for the models trained on a single labeled dataset, FullSup-Cholec80 and FullSup-Each ($D_{1-4}$), with a range of $\sim 13\%$ depending on the training dataset. 
Still, the results for all considered models, including those trained on all 5 labeled datasets are considerably lower than those presented in \ref{tab:r_baseline} when tested on an independent dataset. Visible differences in instruments from all included training centers as well as more subtle variations in acquisition hardware, workflow, etc. may be causative factors. While external evaluations are not yet commonplace in the federated semi-supervised literature \cite{jeong2020federated,zhang2021improving}, this highlights a significant gap that needs to be further investigated and addressed to ensure that clinical applications are sufficiently adaptable to variations (e.g in patient demographics) which may not be adequately represented in available training datasets.}

\begin{table}[ht]
     \centering
     \begin{tabular}{ccc}
        \hline
            & & H$_{\text{External}}$ \\\hline
            \multicolumn{2}{c}{FullSup-Cholec80} & $34.42 \pm 1.87$ \\
            \multirow{4}{*}{FullSup-Each} & $D_1$ & $27.45 \pm 1.71$ \\
            & $D_2$ & $40.40 \pm 3.42$ \\
            & $D_3$ & $33.03 \pm 4.11$ \\
            & $D_4$ & $37.11 \pm 1.81$ \\\hline
            \multicolumn{2}{c}{FedFixMatch} & $34.92 \pm 1.91$ \\
            \multicolumn{2}{c}{FedUDA} & $37.54 \pm 2.61$ \\
            \multicolumn{2}{c}{FedTCC} & \textcolor{blue}{$38.60 \pm 2.70$} \\
            \multicolumn{2}{c}{FedMatch} & $38.12 \pm 1.24$ \\
            \multicolumn{2}{c}{FedRGD} & $37.53 \pm 1.11$ \\
            
            \multicolumn{2}{c}{FedCy (Ours)} & $39.04 \pm 1.63$ \\
            \multicolumn{2}{c}{FedCy-GN (Ours)} & \pmb{$42.03 \pm 1.33$} \\\hline
            \multicolumn{2}{c}{FedAvg} & $42.91 \pm 0.82$ \\
            \multicolumn{2}{c}{FullSup-All} & $46.47 \pm 0.96$ \\\hline
     \end{tabular}
     \caption{F1-score performance on an external client not participating in the training process. We report mean and standard deviation across 3 runs.}
     \label{tab:external}
 \end{table}

\subsection{Effect of Temporal Module}
\label{sec:tecno}
Whereas the previously presented results are based on a ResNet-50 feature extractor, state-of-the-art methods for supervised surgical phase recognition \cite{gao2021trans, czempiel2020tecno} often additionally train temporal modules, such as multi-stage temporal convolutional networks\cite{farha2019ms}, using the learned features extracted from the feature extractor. We study the quality of the features learned by FedCy and the other baselines by training a two-stage temporal convolutional neural network (TeCNO)\cite{czempiel2020tecno}. We use Cholec80 to fine-tune TeCNO for all the semi-supervised baselines, and the corresponding client data for each of the FullSup-Each baselines. For FedAvg and FullSup-All, we correspondingly train TeCNO using federated averaging and a centralized dataset, respectively. In Table \ref{tab:tcn}, we see that the previously noted improvements in the feature extractor (Table \ref{tab:r_baseline}) also translate to gains in performance with the addition of the temporal module. Using the unlabeled data, we see improvements on both Cholec80 and the unlabeled datasets over state-of-the-art phase recognition performance (FullSup-Cholec80). Whereas the substitution of batch normalization with group normalization (FedCy-GN) does not significantly improve the predictive quality on single images (see Table \ref{tab:r_baseline}), when used in tandem with the temporal context of a frame, the learned representational space seems to be much more expressive. \textcolor{blue}{ To ensure a fair comparison, we also reran all the baselines in Table \ref{tab:tcn} with group normalization. We note that FedCy-GN still outperforms all other semi-supervised benchmarks on the unlabeled datasets, by at least 3\% F1-score, while staying on par with the much more label-intensive fully supervised baseline, FullSup-Each. Even without using group normalization, Table \ref{tab:tcn} shows that FedCy outperforms all other benchmarks other than FedRGD, which does use group normalization. To note, even here, FedCy overall performance on all the datasets lies within 1.5 stds of the FedRGD overall performance on all the datasets across the 3 reruns.} Here, FullSup-Cholec80 corresponds to \cite{czempiel2020tecno} but uses the newly generated phase labels and does not use instrument annotations.
 \begin{table}[hb]
     \centering
     \begin{tabular}{cccc}
        \hline
            with TeCNO & Cholec80 & Overall$_{\text{unlabeled}}$ & Overall$_{\text{all}}$ \\\hline
            FullSup-Cholec80 & $78.33 \pm 0.33$&$58.43 \pm 3.37$ & $62.41 \pm 2.75$ \\\hline
            FedFixMatch & $77.92 \pm 1.00$&$57.25 \pm 1.40$ & $61.39 \pm 0.94$ \\
            FedUDA & $78.37 \pm 0.59$&$57.97 \pm 0.92$ & $62.05 \pm 0.68$ \\
            FedTCC & \textcolor{blue}{$78.59 \pm 0.68$} & \textcolor{blue}{$64.05 \pm 1.92$} & \textcolor{blue}{$66.96 \pm 1.67$} \\
            FedMatch & $79.76 \pm 1.00$ & $63.22 \pm 0.73$ & $66.53 \pm 0.52$\\
            FedRGD & \pmb{$81.78 \pm 0.48$} & $66.47 \pm 2.43$ & $69.53 \pm 1.91$\\
            FedCy (Ours) & $79.21 \pm 0.63$& $63.98 \pm 0.47$ & $67.02 \pm 0.50$\\
            FedCy-GN (Ours) & $80.37 \pm 1.29$ & \pmb{$70.09 \pm 0.68$} & \pmb{$72.14 \pm 0.80$}\\\hline
            FullSup-Each & $78.33 \pm 0.33$&$69.32 $ & $71.12$ \\
            FedAvg & $82.28 \pm 1.02$&$76.63 \pm 1.94$ & $77.76 \pm 1.60$ \\
            
            FullSup-All & $81.40 \pm 0.83$&$81.56 \pm 0.91$ & $81.52 \pm 0.79$\\\hline
     \end{tabular}
     \caption{F1-score performance of TeCNO\cite{czempiel2020tecno} model based on feature extractors trained using the presented baselines. We report mean and standard deviation across 3 runs.}
     \label{tab:tcn}
 \end{table}%

\section{Discussion and Conclusion}
\label{sec:conclusion}

We studied the task of surgical phase recognition of laparoscopic cholecystectomy (LC) videos in a federated semi-supervised learning setting, highlighting the feasibility and efficacy of training on unlabeled datasets without the need to explicitly share data. To this end, we proposed a novel FSSL method, FedCy, to efficiently leverage the temporal knowledge found in labeled videos using contrastive learning to guide unsupervised training on unlabeled videos. Comparisons with the state-of-the-art FSSL methods showed significant improvements by our method in the task of surgical phase recognition on labeled, unlabeled and held-out datasets. To conduct this study, we generated a new diverse and large multicenter dataset of LC videos, annotated with 6 phases according to a newly introduced annotation protocol that is robust to workflow variations and reflective of surgical semantics. We believe that such diverse datasets can push multicenter research and evaluation, contributing to the clinical translation of tools for surgical video analysis.

\subsection{\textcolor{blue}{Future Work}}
In the context of our application, several limitations remain that we plan to address in future works. Firstly, the annotation protocol used to annotate the labeled dataset may still not be applicable to hospitals not participating in this study. Both this work and limitation emphasize the need for wider consensus on annotation protocols used to generate the precious datasets that drive work in the field. Besides that, a crucial property of FedCy is the critical role that the labeled dataset plays in guiding the unsupervised training. In this context, using a more representative dataset than Cholec80 \cite{twinanda2016endonet} for supervised training could add another significant performance boost.

Furthermore, in the FSSL scenarios where labeled datasets originate from sources different from those of the unlabeled ones, the data validation split used for hyperparameter tuning and model selection still poses a problem. Most of the FSSL work mentioned in Section \ref{sec:context} simulate the federated learning setup by gathering data, reserving a validation and a test split, then splitting the training data into non-identically distributed datasets. In practice, and in our experiments, data validation splits correspond to only the labeled dataset, and hence hyperparameter tuning and model selection might be biased towards the labeled dataset. Such practical issues could be addressed in future work, for example, by using weak labels.

Finally, future work might investigate how temporal modules can be adapted to FSSL methods, including FedCy. Common practices \cite{czempiel2020tecno, gao2021trans} in the non-federated supervised settings involve using the learned features from the feature extractor as inputs to train a temporal module. Adapting such multi-stage training pipelines in a FSSL setting could boost performance, but is challenging due to memory constraints.

\bibliography{bibliography.bib}

\end{document}